\documentclass[10pt,twocolumn,letterpaper]{article}

\usepackage[pagenumbers]{cvpr} % To force page numbers, e.g. for an arXiv version

\usepackage{graphicx}
\usepackage{amsmath}
\usepackage{amssymb}
\usepackage{booktabs}

\usepackage{algorithm}
\usepackage{algorithmic}
\usepackage{multirow}
\usepackage{bbding}
\usepackage{enumitem}

\usepackage{colortbl} % using color table
\usepackage{xcolor}
\usepackage[pagebackref,breaklinks,colorlinks]{hyperref}

\usepackage[capitalize]{cleveref}
\crefname{section}{Sec.}{Secs.}
\Crefname{section}{Section}{Sections}
\Crefname{table}{Table}{Tables}
\crefname{table}{Tab.}{Tabs.}

\def\eg{\emph{e.g}\onedot} 
\def\ie{\emph{i.e}\onedot}

\newcommand*\samethanks[1][\value{footnote}]{\footnotemark[#1]}

\begin{document}


\title{Augmentation Matters: A Simple-yet-Effective Approach to Semi-supervised Semantic Segmentation}

\author{Zhen Zhao\textsuperscript{\rm 1}\thanks{This work was done during an internship at Baidu VIS.}~~~~Lihe Yang\textsuperscript{\rm 3}~~~~Sifan Long\textsuperscript{\rm 4}~~~~Jimin Pi\textsuperscript{\rm 2}~~~~Luping Zhou\textsuperscript{\rm 1}\thanks{Corresponding authors.}~~~~Jingdong Wang\textsuperscript{\rm 2}\samethanks \\
\textsuperscript{\rm 1}University of Sydney\hspace{16mm}
\textsuperscript{\rm 2}Baidu Inc.\hspace{16mm}
\textsuperscript{\rm 3}Nanjing University\hspace{16mm}
\textsuperscript{\rm 4}Jilin University\\
% {\tt\small \{zhen.zhao, luping.zhou\}@sydney.edu.au~~lihe.yang.cs@gmail.com}\\
% {\tt\small summitlsf@outlook.com~~\{pijimin01, wangjingdong\}@baidu.com}
}
\maketitle

%%%%%%%%% ABSTRACT
\begin{abstract}
    Recent studies on semi-supervised semantic segmentation (SSS) have seen fast progress. Despite their promising performance, current state-of-the-art methods tend to increasingly complex designs at the cost of introducing more network components and additional training procedures. 
    Differently, in this work, we follow a standard teacher-student framework and propose \textbf{AugSeg}, a simple and clean approach that focuses mainly on data perturbations to boost the SSS performance.
    We argue that various data augmentations should be adjusted to better adapt to the semi-supervised scenarios instead of directly applying these techniques from supervised learning.
    Specifically, we adopt a simplified intensity-based augmentation that selects a random number of data transformations with uniformly sampling distortion strengths from a continuous space. 
    Based on the estimated confidence of the model on different unlabeled samples, we also randomly inject labelled information to augment the unlabeled samples in an adaptive manner.
    Without bells and whistles, our simple AugSeg can readily achieve new state-of-the-art performance on SSS benchmarks under different partition protocols\footnote{Code and logs: \url{https://github.com/zhenzhao/AugSeg}.}.
\end{abstract}

%%%%%%%%%%%%%%%%%%%%%%%%%%%%%%%%%%%%
%%  1. introduction
%%%%%%%%%%%%%%%%%%%%%%%%%%%%%%%%%%%%
\section{Introduction}
\label{sec:intro}

\begin{figure}[ht]
  \centering
   \includegraphics[width=0.85\linewidth]{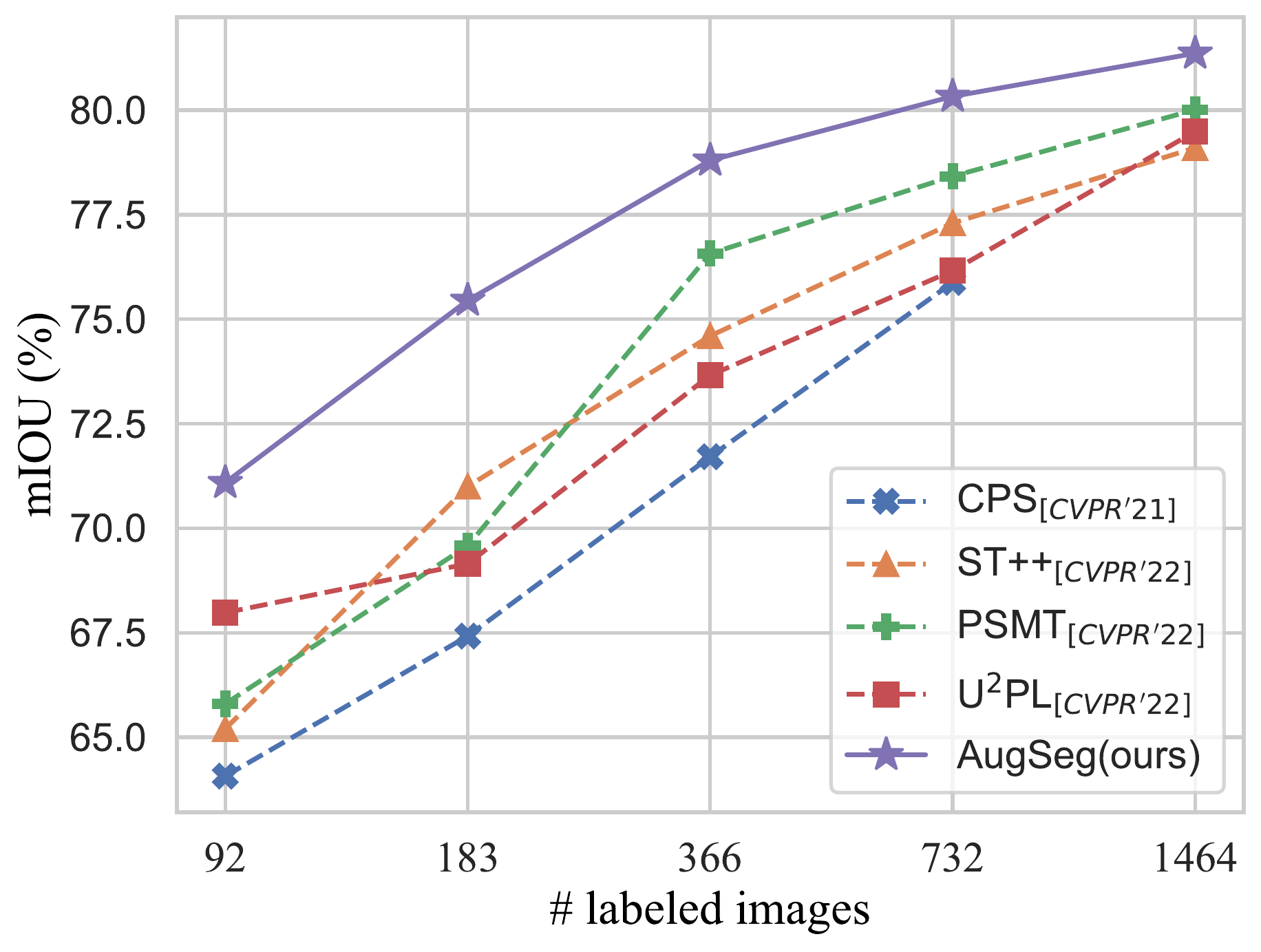}
   \caption{Comparison between current SOTAs and our simple AugSeg on Pascal VOC 2012, using R101 as the encoder.}
   \label{fig:intro}
\end{figure}

Supervised semantic segmentation studies~\cite{seg15fcn,seg17deeplab,seg18deeplabv3plus,zhao2017pyramid} have recently achieved tremendous progress, but their success depends closely on large datasets with high-quality pixel-level annotations. Delicate and dense pixel-level labelling is costly and time-consuming, which becomes a significant bottleneck in practical applications with limited labelled data. To this end, semi-supervised semantic segmentation (SSS)~\cite{sss18advseg,sss19s3gan} has been proposed to train models on less labelled but larger amounts of unlabeled data.

Consistency regularization~\cite{semi1,semi2}, the currently dominant fundamental SSS method, effectively incorporates the training on unlabeled data into standard supervised learning~\cite{sss20cct,sss20dmt}.  It relies on the label-preserving data or model perturbations to produce the prediction disagreement on the same inputs, such that unlabeled samples can be leveraged to train models even if their labeled information is unknown.
Some studies in~\cite{sss21simple,sss20cutmixseg,sss20gct,sss22st++} explored different data augmentations to benefit the SSS training while works in~\cite{ssl17mt,sss20dmt,sss21cps} mainly focused on various model perturbations to obtain competitive SSS performance.
% Many recent attempts~\cite{sss20pseudoseg,sss22PSMT,sss22u2pl} tended to exploit both perturbations.
On top of these fundamental designs, recent state-of-the-art (SOTA) methods aim to integrate extra auxiliary tasks~\cite{sss21c3seg,sss21pc2seg,sss21semisegcontrast,sss22u2pl}, \eg, advanced contrastive learning techniques, and more trainable modules~\cite{sss20ecs,sss20scn,sss22ELN,sss22PSMT}, \eg multiple ensemble models and additional correcting networks, to further improve the SSS performance. 
Despite their promising performance, SSS studies along this line come at the cost of \textbf{requiring more complex methods}, \eg, extra network components or additional training procedures.

\begin{table}
\centering
\resizebox{0.49\textwidth}{!}{
\begin{tabular}{r|cc|ccc|ccc}
\toprule
\multirow{2}{*}{Method} & \multicolumn{2}{c|}{Augmentations}& \multicolumn{3}{c|}{More Supervision} & \multicolumn{3}{c}{Pseudo-rectifying} \\
\cmidrule(lr){2-3} \cmidrule(lr){4-6} \cmidrule(lr){7-9}
& SDA & FT & MBSL & CT & UCL & UAFS & ACN & PR \\
\midrule
CCT~\cite{sss20cct} &  & \checkmark & \checkmark & \checkmark &  &  &  &  \\
% SCN~\cite{sss20scn} & - & - & - & \checkmark & - & - & - & \checkmark & -\\
ECS~\cite{sss20ecs} &  &  & \checkmark &  &  &  & \checkmark & \\
SSMT~\cite{sss22ssmt}  & \checkmark &  & \checkmark &  &  & \checkmark &  & \\
PseudoSeg~\cite{sss20pseudoseg} & \checkmark &  &  &  &  & \checkmark & & \\
CAC~\cite{sss21cac} &  &  & \checkmark &  & \checkmark & \checkmark &  & \\
DARS~\cite{sss21dars} & \checkmark &  & \checkmark &  &  &  &  & \checkmark\\
AEL~\cite{sss21ael} & \checkmark &  &  &  &  &  &  & \checkmark\\
PC${}^2$Seg~\cite{sss21pc2seg} & \checkmark &  & \checkmark &  & \checkmark & \checkmark &  & \\
C3-Semiseg~\cite{sss21c3seg} & \checkmark &  &  &  & \checkmark & \checkmark &  & \checkmark \\
SimpleBase~\cite{sss21simple} & \checkmark &  & \checkmark &  &  & \checkmark &  & \\
ReCo~\cite{sss22reco} & \checkmark &  &  &  & \checkmark & \checkmark &  & \\
CPS~\cite{sss21cps} & \checkmark &  &  & \checkmark &  &  &  & \\
ST++~\cite{sss22st++} & \checkmark &  & \checkmark &  &  &  &  & \\
ELN~\cite{sss22ELN} & \checkmark &  & \checkmark &  &  &  & \checkmark & \\
USRN~\cite{sss22USR} & \checkmark &  & \checkmark &  &  & \checkmark &  & \checkmark\\
PSMT~\cite{sss22PSMT} & \checkmark & \checkmark & \checkmark &  &  & \checkmark &  & \\
U$^2$PL~\cite{sss22u2pl} & \checkmark &  &  &  & \checkmark & \checkmark &  & \checkmark\\
\midrule
\textbf{AugSeg (ours)} & \checkmark &  &  &  &  &  &  &  \\
\bottomrule
\end{tabular}
}
\caption{ Comparison of recent SSS algorithms in terms of ``Augmentations", ``More supervision", and ``Pseudo-rectifying" (sorted by their publication date). We explain the abbreviations as follows. ``\textbf{SDA}": Strong data augmentations, including various intensity-based and cutmix-related augmentations, ``\textbf{FT}": Feature-based augmentations, ``\textbf{MBSL}": multiple branches, training stages, or losses, ``\textbf{CT}": Co-training, ``\textbf{UCL}": unsupervised contrastive learning, ``\textbf{UAFS}": uncertainty/attention filtering/sampling, ``\textbf{ACN}": additional correcting networks, ``\textbf{PR}": prior-based re-balancing techniques. \textbf{Note that}, branches of ``more supervision" and ``pseudo-rectifying" typically require more training efforts. Differently, our method enjoys the best simplicity but the highest performance.
}
\label{tab:sss:cmps}
\end{table}

In this paper, we break the trend of recent SOTAs that combine increasingly complex techniques and propose \textbf{AugSeg}, a simple-yet-effective method that focuses mainly on data perturbations to boost the SSS performance. 
Although various auto data augmentations~\cite{cubuk2018autoaugment,augs20randaugment} and cutmix-related transformations~\cite{augs19cutmix,sss20cutmixseg} in supervised learning have been extensively utilized in previous SSS studies, we argue that these augmentations should be adjusted precisely to better adapt the semi-supervised training. 
\textbf{On one hand}, these widely-adopted auto augmentations in existing SSS studies aim to search the optimal augmentation strategies from a predefined finite discrete space. 
However, data perturbations in semi-supervised learning consist in generating prediction disagreement on the same inputs, without a specific ``optimal" objective or a pre-defined discrete searching space. 
Thus, we simplify existing randomAug~\cite{augs20randaugment} and design a highly random intensity-based augmentation, which selects a random number of different intensity-based augmentations and a random distortion strength from a continuous space. 
\textbf{On the other hand}, random copy-paste~\cite{augs21copy} among different unlabeled samples can yield effective data perturbations in SSS, but their mixing between corresponding pseudo-labels can inevitably introduce confirmation bias~\cite{ssl20bias}, especially on these instances with less confident predictions of the model. Considering the utilization efficiency of unlabeled data, we simply mix labeled samples with these less confident unlabeled samples in a random and adaptive manner, \ie, adaptively injecting labeled information to stabilize the training on unlabeled data.
Benefiting from the simply random and collaborative designs, AugSeg requires no extra operations to handle the distribution issues, as discussed in \cite{sss21simple}.

Despite its simplicity, AugSeg obtains new SOTA performance on popular SSS benchmarks under various partition protocols. As shown in \Cref{fig:intro}, AugSeg can consistently outperform current SOTA methods by a large margin. For example, AugSeg achieves a high mean intersection-over-union (mIoU) of 75.45\% on classic Pascal VOC 2012 using only 183 labels compared to the supervised baseline of 59.10\% and previous SOTA of 71.0\% in \cite{sss22st++}. We attribute these remarkable performance gains to our revision -- that various data augmentations are simplified and adjusted to better adapt to the semi-supervised scenarios. Our main contributions are summarized as follows,
\begin{itemize}
    \item We break the trend of SSS studies that integrate increasingly complex designs and propose AugSeg, a standard and simple two-branch teacher-student method that can achieve readily better performance.
    \item We simply revise the widely-adopted data augmentations to better adapt to SSS tasks by injecting labeled information adaptively and simplifying the standard RandomAug with a highly random design.
    \item We provide a simple yet strong baseline for future SSS studies. Extensive experiments and ablations studies are conducted to demonstrate its effectiveness. 
\end{itemize}

%%%%%%%%%%%%%%%%%%%%%%%%%%%%%%%%%%%%
%%  2. related works
%%%%%%%%%%%%%%%%%%%%%%%%%%%%%%%%%%%%

\begin{figure*}
    \centering
    \includegraphics[width=0.832\textwidth]{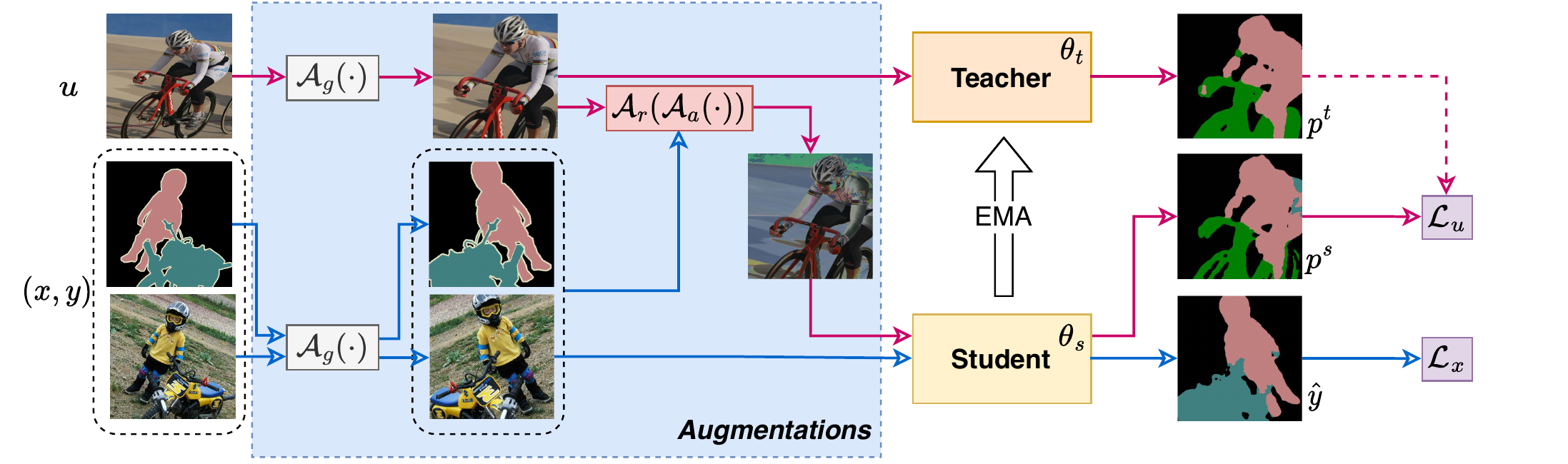}
    \caption{Diagram of AugSeg. In a standard teacher-student framework, AugSeg trains the student model, parameterized by $\theta_s$, on labeled data $(x,y)$ and unlabeled data $u$ simultaneously, via minimizing the corresponding supervised loss $\mathcal{L}_x$ and unsupervised consistency loss $\mathcal{L}_u$, respectively. The teacher model, parameterized by $\theta_t$, is updated by the exponential moving averaging (EMA) of $\theta_s$, and generates the pseudo-label on unlabeled data, $p^t$. The core of AugSeg is to apply various augmentation techniques on input unlabeled samples, including the weak geometrical augmentation $\mathcal{A}_g$, the random intensity-based augmentation $\mathcal{A}_r$ and the adaptive label-injecting augmentation $\mathcal{A}_a$. The red and blue lines represent the forward path of labeled and unlabeled data, respectively. The dashed line means``stop gradient".
    }
    \label{fig:diagram}
\end{figure*}

\section{Related work}
\label{sec:rwork}

The key to semi-supervised learning lies in effectively leveraging the unlabeled data~\cite{semi1,ssl05entropy,ssl13pseudo}. Recent consistency regularization (CR)~\cite{ssl17mt,ssl17Pi} has become a fundamental semi-supervised technique to train models on labeled and unlabeled data simultaneously. Such CR-based methods, in either classification~\cite{ssl20uda,ssl20fixmatch} or segmentation tasks~\cite{sss20cutmixseg,sss20pseudoseg}, rely on various perturbation techniques to generate disagreement on the same inputs, so that models can be trained by enforcing prediction consistency on unlabeled data without knowing labeled information. Along this line, many SSS methods have been proposed recently. 

Based on our summary, as shown in \Cref{tab:sss:cmps}, there are three different main directions to enhance the SSS performance, including ``augmentations", ``more supervision," and ``pseudo-rectifying". Almost all existing studies applied various strong data augmentations to perturb unlabeled data while some of them~\cite{sss20cct,sss22PSMT} also perturbed the inputs at the feature level.
In the branch of ``more supervision", multiple training branches, training stages, or losses (MBSL) are widely adopted from the perspective of model perturbations~\cite{sss22ssmt,sss22st++,sss22PSMT,sss21cac}. As the quality of pseudo-labels is critical for semi-supervised training~\cite{ssl22lassl}, ECS~\cite{sss20ecs} and ELN~\cite{sss22ELN} also introduced additional trainable correcting networks (ACN) to further polish the pseudo-labels.
Recent SOTA methods~\cite{sss22u2pl,sss22PSMT,sss22USR,sss22ELN} can achieve promising performance at the cost of combining increasingly complex mechanisms, \eg contrastive learning~\cite{self1} and multiple ensembling models. Differently, in this work, we aim to propose a simple and clean method that relies mainly on data augmentations to boost the SSS performance.

As the most straightforward and effective way to produce label-preserving perturbations, data augmentations have played a central role in CR-based semi-supervised studies~\cite{ssl20uda,ssl18vat,sss20cutmixseg}. Recently, various auto augmentation strategies~\cite{2020Randaugment,cubuk2018autoaugment,trivialaug} in supervised learning have been widely adopted in semi-supervised research.
However, directly applying such techniques is not satisfying for the following reasons. The goal of auto augmentations is to find out the optimal augmentation strategies. Such studies aim to search best augmentation operations and distortion strengths from a finite discrete space. In contrast, the objective of applying data augmentation in SSS is to generate different inputs without specific optimal goals and searching spaces. Besides, as discussed in \cite{sss21simple}, directly applying these augmentations may over-distort unlabeled data and hurt the data distribution, resulting in  performance degradation. Instead of using additional rectifying strategies like distribution-specific BN~\cite{chang2019domain}, we simplify the standard randomAug~\cite{augs20randaugment} with a highly random design. Instead of using a predefined number of augmentations with finite and discrete strength possibilities, we select a random number of augmentations and sample the augmentation strength uniformly from a continuous interval. In this way, our design enjoys better data diversity and is less likely to over-distort samples. 
% We also inject the labelled information adaptively to stabilize the training on unlabeled samples further.

%%%%%%%%%%%%%%%%%%%%%%%%%%%%%%%%%%%%
%%  3. methods
%%%%%%%%%%%%%%%%%%%%%%%%%%%%%%%%%%%%
\section{Augseg}
\label{sec:method}

In this section, we first present an overview of our simple AugSeg and then describe our main augmentation strategies, random intensity-based and adaptive cutmix-based augmentations, in \Cref{sec:method:intensity} and \Cref{sec:method:adaptive}, respectively.

\subsection{Overview}
\label{sec:method:overview}

Following current dominant consistency regularization methods in semi-supervised learning~\cite{semi2},  we train our segmentation model on labeled and unlabeled data simultaneously. As shown in \Cref{fig:diagram}, we adopt a simple and clean semi-supervised framework, which consists of a student model and a teacher model, parameterized by $\theta_s$ and $\theta_t$, respectively. 
Specifically, the teacher model is capable of producing pseudo-labels for training on unlabeled data, and will be updated gradually via the exponential moving averaging of the student weights, \ie,
\begin{align}
    \theta_t \leftarrow \alpha \theta_t + (1 - \alpha) \theta_s,
\end{align}
where $\alpha$ is a common momentum parameter, which is, following \cite{ssl17mt}, set as 0.999 by default. On the other hand, at each iteration, provided with a batch of labeled samples $\mathcal{B}_x=\{(x_i, y_i)\}_{i=1}^{|\mathcal{B}_x|}$ and a batch of unlabeled samples $\mathcal{B}_u=\{u_i\}_{i=1}^{|\mathcal{B}_u|}$, we aim to train the student model via minimizing a supervised loss $\mathcal{L}_x$ and an unsupervised consistency loss $\mathcal{L}_u$ at the same time. Thus the total training loss for the student model is,
\begin{align}
    \label{equ:loss:total}
    \mathcal{L} = \mathcal{L}_x + \lambda_u \mathcal{L}_u,
\end{align}
where $\lambda_u$ is a scalar hyper-parameter to adjust the unsupervised loss weight. Similar to most SSS methods~\cite{sss18advseg,sss20cct,sss20ecs}, we adopt a standard pixel-wise cross-entropy loss $\ell_{ce}$ to train on labeled data directly, 
\begin{align}
    \mathcal{L}_x = \frac{1}{|\mathcal{B}_x|} \sum_{i=1}^{|\mathcal{B}_x|} \frac{1}{H\times W}\sum_{j=1}^{H\times W} \ell_{ce}(\hat{y}_i(j), y_i(j)),
\end{align}
where $\hat{y}_i\!=\!f(\mathcal{A}_g(x_i); \theta_s)$, represents the segmentation result of the student model on the $i$-th weakly-augmented labeled instance. 
$j$ represents the $j$-th pixel on the image or the corresponding segmentation mask with a resolution of $H\times W$. The weak geometrical augmentation $\mathcal{A}_g$, as shown in \Cref{tab:augs}, includes standard resizing, cropping, and flipping operations. As for leveraging the unlabeled data, which is the key to semi-supervised learning, we rely mainly on the data perturbation $\mathcal{T(\cdot)}$ to generate the prediction disagreement. First, we obtain the segmentation predictions, $p_i^s$ and $p_i^t$, of the student model on augmented $\mathcal{T}(u_i)$ and of the teacher model on augmented $\mathcal{A}_g(u_i)$, respectively, 
\begin{align}
    p^t_i &= f (\mathcal{A}_g(u_i); \theta_s), \\
    p^s_i &= f (\mathcal{T}(\mathcal{A}_g(u_i)); \theta_t).
\end{align}
Subsequently, the unlabeled loss is formulated as,
\begin{align}
    \mathcal{L}_{u} = \frac{1}{|\mathcal{B}_u|}\!\sum_{i=1}^{|\mathcal{B}_u|}\!\frac{1}{H\times W}\sum_{j=1}^{H\!\times\!W}\!\ell_{ce}(p_i^s(j), p^t_i(j)).
\end{align}
Different from recent SSS methods, our AugSeg follows a clean and simple two-branch teacher-student framework. We rely mainly on our augmentation strategy $\mathcal{T}(\cdot)$ to produce prediction disagreement on the same input, which is also the key to semi-supervised learning. The augmentation $\mathcal{T}$, the core of AugSeg, consists of two kinds of augmentation in a cascade fashion, i.e., $\mathcal{T}(\cdot) = \mathcal{A}_r(\mathcal{A}_a(\cdot))$, which are detailed in following sections.

\begin{table}[t]
\centering
\resizebox{0.48\textwidth}{!}{
\begin{tabular}{ll}
\toprule
% \multicolumn{2}{c}{\textbf{Weak Geometrical Augmentation} - Apply all} \\
\multicolumn{2}{c}{\textbf{Weak Geometrical Augmentation} - Apply all} \\
\midrule
Random  Scale & Randomly resizes the image by $[0.5, 2.0]$. \\
Random Flip & Horizontally flips the image with a probability of 0.5. \\
Random Crop & Randomly crops an region from the image. \\
\midrule
% \midrule
\multicolumn{2}{c}{\textbf{Random Intensity-based Augmentation} - Apply $k$ randomly } \\
\midrule
Identity & Returns the original image. \\
Autocontrast & Maximizes (normalize) the image contrast.\\
Equalize & Equalize the image histogram.\\
Gaussian blur & Blurs the image with a Gaussian kernel.\\
Contrast & Adjusts the contrast of the image by [0.05, 0.95]. \\
Sharpness & Adjusts the sharpness of the image by [0.05, 0.95]. \\
Color & Enhances the color balance of the image by [0.05, 0.95] \\
Brightness & Adjusts the brightness of the image by [0.05, 0.95] \\
Hue & Jitters the hue of the image by [0.0, 0.5] \\
Posterize & Reduces each pixel to [4,8] bits.\\
Solarize & Inverts image pixels above a threshold from [1,256).\\
\bottomrule
\end{tabular}
}
\caption{List of various image transformations in the weak geometrical augmentation and random intensity-based augmentation.}
\label{tab:augs}
\end{table}

\subsection{Random Intensity-based Augmentations}
\label{sec:method:intensity}

\begin{figure}
    \centering
    \includegraphics[width=0.95\linewidth]{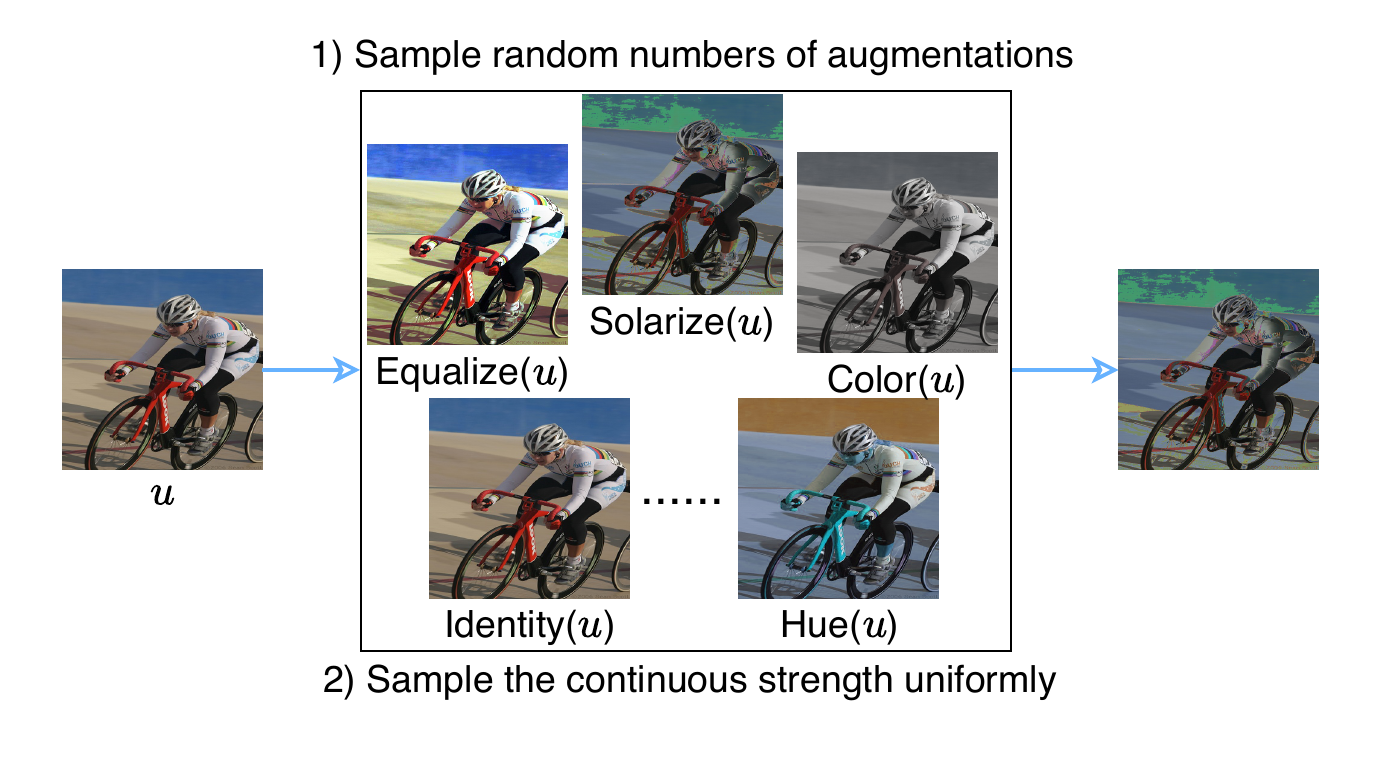}
    \caption{A visualization of random intensity-based augmentation.}
    \label{fig:aug:intensity}
\end{figure}

In most existing semi-supervised learning studies, either in classification tasks or segmentation tasks, various auto augmentation techniques~\cite{cubuk2018autoaugment}, especially the simplified RandomAug~\cite{augs20randaugment}, have been widely adopted to perturb unlabeled samples. However, its different objective from semi-supervised learning limits its effectiveness.
Specifically, the goal of various auto augmentations is to search for the optimal augmentation strategies for a specific downstream task. RandomAug further simplified this searching procedure in a finite discrete space. Whereas, the goal of data perturbation in semi-supervised learning is to generate two different views from the same image, where no specific optimal augmentation strategy is required. Besides, as discussed in \cite{sss21simple}, over-distorted augmentations will hurt the data distribution and degrade the SSS performance. To this end, we design a random intensity-based augmentation, denoted by $\mathcal{A}_r$, to perturb unlabeled data. As shown in \cref{fig:aug:intensity}, we 
\begin{itemize}
    \item sample the distorting degree uniformly in a continuous space instead of a finite discrete space. 
    \item sample a random number of augmentations, bounded by a maximum value of $k$, from an augmentation pool instead of using a fixed number.
    \item remove strong intensity-based transformations like the Invert operations~\cite{sss21simple} in our augmentation pool. Our pool is directly simplified from the pool in RandomAug~\cite{2020Randaugment}, as shown in \Cref{tab:augs}.
\end{itemize}
In this way, our random intensity-based augmentation can enjoy better data diversity and adapt more to tsemi-supervised tasks. More importantly, different from \cite{sss21simple}, our highly random designs will not hurt the data distribution remarkably. Thus we can get rid of additional distribution-specific revisions~\cite{chang2019domain} and extra filtering strategies~\cite{sss21simple}.

\subsection{Adaptive CutMix-based augmentations}
\label{sec:method:adaptive}

\begin{table*}[ht]
\centering
\begin{tabular}{lcccccc}
\toprule
Method & Encoder &1/16 (92) &1/8 (183) &1/4 (366) &1/2 (732) &Full (1464) \\
\hline
Supervised &R50            & 44.03 & 52.26 &  61.65 & 66.72 & 72.94 \\
PseudoSeg~\cite{sss20pseudoseg}    &R50   & 54.89 & 61.88 & 64.85 & 70.42 & 71.00 \\
PC$^2$Seg~\cite{sss21pc2seg} &R50       & 56.90 & 64.63 & 67.62 & 70.90 & 72.26 \\
% \rowcolor{blue!30} 
\textbf{AugSeg} 
&R50 & \cellcolor{blue!30}64.22 & \cellcolor{blue!30}72.17 & \cellcolor{blue!30}76.17 & \cellcolor{blue!30}77.40 & \cellcolor{blue!30}78.82 \\
\midrule
Supervised &R101            & 43.92 & 59.10 & 65.88 & 70.87 & 74.97 \\
CutMix-Seg~\cite{sss20cutmixseg}    &R101       & 52.16 & 63.47 & 69.46 & 73.73 & 76.54 \\ 
PseudoSeg~\cite{sss20pseudoseg}    &R101   & 57.60 & 65.50 & 69.14 & 72.41 & 73.23 \\
PC$^2$Seg~\cite{sss21pc2seg} &R101        & 57.00 & 66.28 & 69.78 & 73.05 & 74.15 \\
CPS~\cite{sss21cps} &R101 & 64.07 & 67.42 & 71.71 & 75.88 & - \\
PS-MT~\cite{sss22PSMT} % $^\dag$  
&R101                  & 65.80 & 69.58 & 76.57 &78.42 & 80.01 \\
ST++~\cite{sss22st++} % $^\dag$  
&R101            & 65.20 & 71.00 & 74.60 & 77.30 & 79.10 \\
U$^2$PL~\cite{sss22u2pl} %$^\dag$  
&R101          & 67.98 & 69.15 & 73.66 & 76.16 & 79.49 \\
% \rowcolor{red!30} 
\textbf{AugSeg}
&R101 & \cellcolor{blue!30}71.09 & \cellcolor{blue!30}75.45 & \cellcolor{blue!30}78.80 & \cellcolor{blue!30}80.33 & \cellcolor{blue!30}81.36 \\
\bottomrule
\end{tabular}
\caption{Compared with the state-of-the-art methods on classic Pascal VOC 2012 val set under different partition protocols. `1/n'  means  that `1/n' data is used as labeled dataset, and the remaining images are used as unlabeled dataset.} 
% Best results on R50 and R101 are highlighted in \textcolor{blue!30}{blue} and \textcolor{red!30}{red}, respectively.}
% $\dag$ means more unlabeled dataset with a total amount of 10582 images, i.e. augmented by the SBD dataset, are used for training.}
\label{tab:voc:fine}
\end{table*}

\begin{figure}[t]
    \centering
    \includegraphics[width=0.98\linewidth]{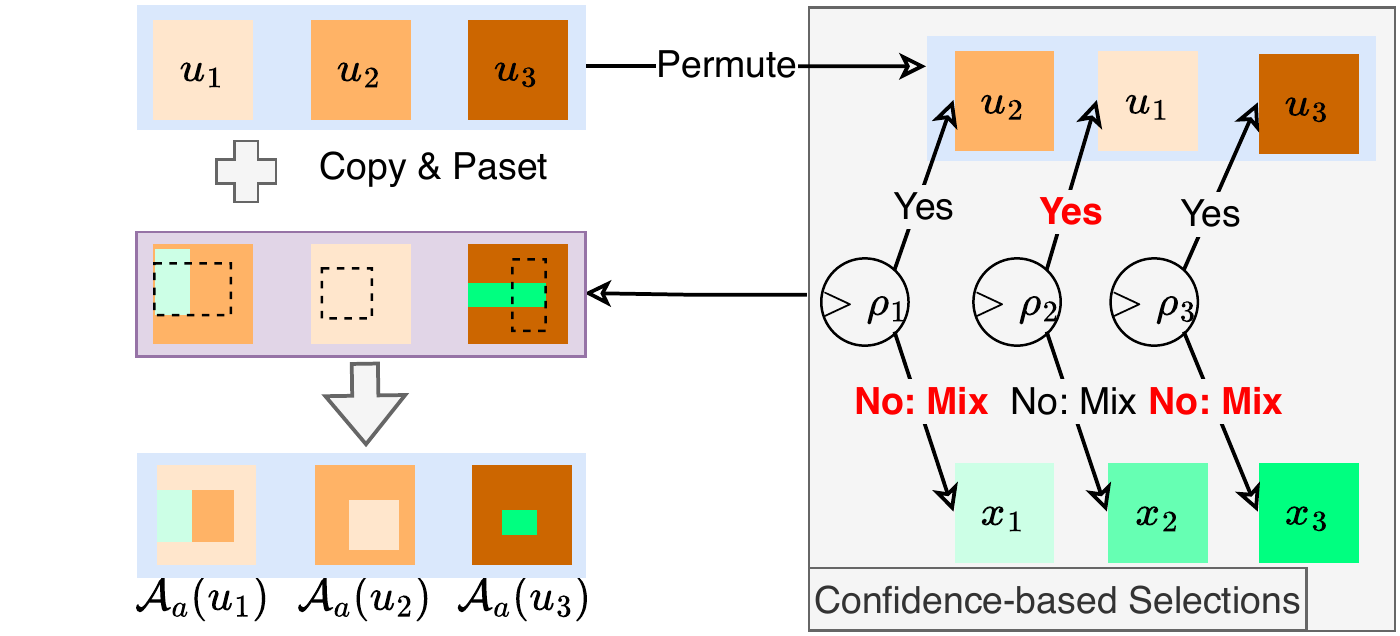}
    \caption{A visualization of adaptive label-injecting CutMix augmentation in a mini-batch. $x_i$ and $u_i$ denote the labeled and unlabeled crops, respectively. $\rho_i$ denote the confidence score for $i$-th unlabeled sample. The core idea of $\mathcal{A}_a$ is that, these less confident unlabeled samples, with lower values of $\rho_i$, are more likely to be aided (mixed) by these confident labeled samples.}
    \label{fig:aug:cut}
\end{figure}

CutMix-related~\cite{augs19cutmix, augs17cutout} or copy-paste~\cite{augs21copy} augmentations have shown their effectiveness in supervised and semi-supervised segmentation tasks. Recent studies in SSS \cite{sss20cutmixseg,sss22u2pl,sss22PSMT} apply the random copy-paste between unlabeled samples within a mini-batch and revise their pseudo-label accordingly. However, relying highly on the pseudo-labels may inevitably result in confirmation bias~\cite{bias2019}, especially for some difficult-to-train samples, or at the early training stages. Thus we tend to leverage the confident labeled samples to augment unlabeled data, so that labeled information can be fully exploited. However, mixing confident labeled information to unlabeled data is naturally beneficial but may under-utilize the unlabeled data. It is simply because some regions of unlabeled data are covered by regions from labeled samples and never utilized during the training. To this end, as shown in \Cref{fig:aug:cut}, we design an adaptive label-injecting augmentation that can make full use of labeled data to aid the training on unlabeled samples in an \textbf{instance-specific and confidence-adaptive} manner. In specific, we first estimate a confidence score, $\rho_i$, indicating the confidence level of the current model on its prediction on $i$-th unlabeled instance,
\begin{equation}
        \rho_i\!=\!\frac{1}{H\!\times\!W}\sum_{j=1}^{H\! \times\!W} \max(p^t_i(j)) (1 - \frac{- \sum p^t_i(j) \log p^t_i(j)}{\log N})
\end{equation}
where we use the weighted average of the normalized prediction entropy on $u_i$ to estimate the confidence score. Apparently, the score $\rho$ is instance-specific and closely related to the generalization probability of the current model. We then use $\rho_i$ as a triggering probability to randomly apply the mixing between labeled and unlabeled instances to obtain the mixing candidates $\{u_n^{'}\}$,
% \begin{align}
% \left\{
% \begin{aligned}
%     u_n^{'} &\leftarrow M_n \odot u_n + (\mathbf{1} - M_n) \odot x_n   \\
%     p^{t'}_n &\leftarrow M_n \odot p^t_n + (\mathbf{1} - M_n) \odot y, 
% \end{aligned}\right.
% \end{align}
\begin{align}
    u_n^{'} &\leftarrow M_n \odot u_n + (\mathbf{1} - M_n) \odot x_n,
\end{align}
where $M_n$ denotes the randomly generated region
mask. After that, we apply the final mixing step between unlabeled instance $\{u_m\}$ and the permuted mixing candidates $\{u_n^{'}\}$,
\begin{align}
    \mathcal{A}_a(u_m) &\leftarrow M_m \odot u_m + (\mathbf{1} - M_m) \odot u_n^{'},
\end{align}
where $M_m$, similar to $M_n$, denotes the randomly generated binary region mask.

%%%%%%%%%%%%%%%%%%%%%%%%%%%%%%%%%%%%
%%  4. Experiments
%%%%%%%%%%%%%%%%%%%%%%%%%%%%%%%%%%%%
\begin{table*}[ht]
  \centering
  % \begin{tabular}{@{}lrrrrrrrr@{}}
  \begin{tabular}{lcccccc}
  \toprule
  \multirow{2}{*}{Method} & 
  \multicolumn{3}{c}{ResNet-50} & 
  \multicolumn{3}{c}{ResNet-101} \\ 
  \cmidrule(lr){2-4} \cmidrule(lr){5-7}
  & {1/16 (662)} & {1/8 (1323)} & {1/4 (2646)} &{1/16 (662)} & {1/8 (1323)} & {1/4 (2646)} \\ \midrule
  Supervised
  & 63.72 & 68.49 & 72.46  
  & 67.76 & 72.13 & 75.04  \\
  MT~\cite{ssl17mt} 
  & 66.77 & 70.78 & 73.22 
  & 70.59 & 73.20 & 76.62  \\
  CCT~\cite{sss20cct} 
  & 65.22 & 70.87 & 73.43 
  & 67.94 & 73.00 & 76.17 \\
  GCT~\cite{sss20gct} 
  & 64.05 & 70.47 & 73.45 
  & 69.77 & 73.30 & 75.25 \\
  CPS \cite{sss21cps} 
  & 68.21 & 73.20 & 74.24 
  & 72.18 & 75.83 & 77.55  \\
  CPS w/ CutMix \cite{sss21cps} & 71.98 & 73.67 & 74.90 & 74.48 & 76.44 & 77.68 \\
  ST++~\cite{sss22st++}
  & 72.60 & 74.40 & 75.40 
  & 74.50 & 76.30 & 76.60  \\
  % \rowcolor{green!30} 
  PS-MT  \cite{sss22PSMT}
  & 72.83 & 75.70 & 76.43 
  & 75.50 & \cellcolor{blue!30}78.20 & 78.72  \\
  \textbf{AugSeg}
  & \cellcolor{blue!30}74.66 & \cellcolor{blue!30}75.99 & \cellcolor{blue!30}77.16 
  & \cellcolor{blue!30}77.01 & 77.31 & \cellcolor{blue!30}78.82 \\
  \midrule
  Supervised$^\ddag$
  & 67.66 & 71.91 & 74.53 
  & 70.63 & 75.02 & 76.47 \\
  U$^2$PL$^\ddag *$~\cite{sss22u2pl}     & 74.74 & 77.44 & 77.51 
  & 77.21 & 79.01 & 79.30 \\
  \textbf{AugSeg}$^\ddag$ 
  & \cellcolor{blue!30}77.28 & \cellcolor{blue!30}78.27 & \cellcolor{blue!30}78.24 
  & \cellcolor{blue!30}79.29 & \cellcolor{blue!30}81.46 & \cellcolor{blue!30}80.50 \\
  \bottomrule
  \end{tabular}
  \caption{Comparison with the state-of-the-art on the PASCAL VOCAug val set under different partition protocols. The VOCAug trainset consists of 10,582 labeled samples in total. $\ddag$ means the same split as U$^2$PL, which prioritizes selecting high-quality labels from classic VOCs. Other methods use the same split as CPS. $*$ presents our reproduced results for U$^2$PL~\cite{sss22u2pl} using ResNet-50.}
  \label{tab:voc:blender}
\end{table*}

\section{Experiments}
\label{sec:exps}
In this section, we first describe our experimental setups and then compare our method with recent SOTAs on SSS benchmarks. A series of ablation studies are also conducted to verify the effectiveness and stability of AugSeg further.

\subsection{Experimental setup}
\noindent\textbf{Dataset}. We examine the efficacy of AugSeg on two widely used segmentation datasets, Pascal VOC 2012~\cite{data15voc} and Cityscapes~\cite{data16citys}. Pascal VOC 2012 is a standard semantic segmentation benchmark with 21 semantic classes (including the background). The classic VOC 2012 includes 1,464 fine-labeled training images and 1,449 validating images. Following~\cite{sss21cps,sss22st++,sss22u2pl}, we also include the additional coarsely-labeled 9,118 images from the Segmentation Boundary dataset (SBD)~\cite{data11vocaugs} as the training images, leading to 10,582 images in total. We adopt the same partition protocols in~\cite{sss21cps,sss22u2pl} to evaluate our method on both \texttt{classic} and \texttt{blender} sets. Cityscapes consists of 19 semantic classes of urban scenes, including 2,975 training and 500 validating images with fine annotations.

\noindent
\textbf{Training}. Following previous SSS studies~\cite{sss22ELN,sss22u2pl,sss22PSMT}, we adopt DeepLabV3$+$~\cite{seg18deeplabv3plus} with ResNet~\cite{he16resnet} pretrained on ImageNet~\cite{data09imagenet} as our segmentation backbone. Different from U$^2$PL~\cite{sss22u2pl}, we use an output stride of 16 by default (instead of using 8). We use an SGD optimizer with a momentum of 0.9 and a polynomial learning-rate decay with an initial value of 0.01 to train the student model. Referring to \cite{sss21cps,sss22u2pl},  we use the crop size of $512\times512$ with a training epoch of 80 and the crop size of $800\times800$ with a training epoch of 240 on VOC2012 and Cityscapes, respectively. A  batch size of 16 and the sync-BN are adopted for both datasets.

% \noindent
\textbf{Evaluations}. We use the mean of intersection-over-union to evaluate the segmentation performance for all runs, using ResNet-50 and ResNet-101 as the encoder separately. Following CPS~\cite{sss21cps} and U$^2$PL~\cite{sss22u2pl}, we also adopt the sliding evaluation to examine the performance on validation images of Cityscapes with a resolution of $1024\times2048$.

\begin{table*}[ht]
  \centering
  % \begin{tabular}{@{}lrrrrrrrr@{}}
  \begin{tabular}{lcccccccc}
  \toprule
  \multirow{2}{*}{Method} & 
  \multicolumn{4}{c}{ResNet-50} & \multicolumn{4}{c}{ResNet-101} \\ 
  \cmidrule(lr){2-5} \cmidrule(lr){6-9}
   & {1/16(186)} &
  {1/8(372)} &
  {1/4(744)} &
  {1/2(1488)} &
  {1/16(186)} &
  {1/8(372)} &
  {1/4(744)} &
  {1/2(1488)}  \\ 
  \midrule
  Supervised
  & 63.34 & 68.73 & 74.14 & 76.62 
  & 64.77 & 71.64 & 75.24 & 78.03 \\
  MT \cite{ssl17mt} 
  & 66.14 & 72.03 & 74.47 & 77.43 
  & 68.08 & 73.71 & 76.53 & 78.59 \\
  CCT \cite{sss20cct} 
  & 66.35 & 72.46 & 75.68 & 76.78 
  & 69.64 & 74.48 & 76.35 & 78.29 \\
  GCT \cite{sss20gct} 
  & 65.81 & 71.33 & 75.30 & 77.09 
  & 66.90 & 72.96 & 76.45 & 78.58 \\
  CPS \cite{sss21cps} 
  & 69.79 & 74.39 & 76.85 & 78.64 
  & 70.50 & 75.71 & 77.41 & 80.08 \\
  % CPS w/ CutMix \cite{sss21cps} & 74.47 & 76.61 & 77.83 & 78.77 & 74.72 & 77.62 & 79.21 & 80.21 \\
  CPS $^{*}$ \cite{sss21cps} & - & - & - & - & 69.78 &  74.31 &74.58 & 76.81 \\
  PS-MT\dag  \cite{sss22PSMT}
  & - & 75.76 & 76.92 & 77.64 
  & - & 76.89 & 77.60 & 79.09 \\
  U$^2$PL~\cite{sss22u2pl} 
  & 69.03 & 73.02 & 76.31 & 78.64
  & 70.30 & 74.37 & 76.47 & 79.05 \\
  \midrule
  \textbf{AugSeg} 
  & \cellcolor{blue!30}73.73 & \cellcolor{blue!30}76.49 & \cellcolor{blue!30}78.76 & \cellcolor{blue!30}79.33 & \cellcolor{blue!30}75.22 & \cellcolor{blue!30}77.82 & \cellcolor{blue!30}79.56 & \cellcolor{blue!30}80.43  \\
  \bottomrule
  \end{tabular}
  \caption{Comparison with state-of-the-art on Cityscapes val set under different partition protocols. Cityscapes (Citys) includes 2, 975 samples in total. $*$ means the reproduced results in U$^2$PL~\cite{sss22u2pl}. All the results are reported by the sliding evaluations. \dag means PS-MT~\cite{sss22PSMT} runs more epochs (320, 450, 550 epochs on 1/8, 1/4, 1/2 splits, respectively) than ours (240 epochs for all splits.)}
  \label{tab:citys}
\end{table*}

\begin{table}[ht]
\centering
% \resizebox{0.9\linewidth}{!}{
\begin{tabular}{ccc|cc}
\toprule
  \multicolumn{3}{c}{AugSeg} & \multicolumn{2}{c}{mIoU} \\ 
  \cline{1-3} \cline{4-5}
MT & $\mathcal{A}_r$ & $\mathcal{A}_a$ &  VOC (366) & Citys (744)\\
\hline
 &  & & 61.65 \footnotesize{(\textcolor{blue}{supervised})} & 74.14 \footnotesize{(\textcolor{blue}{supervised})}\\
\checkmark &  & & 69.06 \footnotesize{(\textcolor{blue}{7.41$\uparrow$})} & 75.96  \footnotesize{(\textcolor{blue}{1.82$\uparrow$})}\\
\checkmark &\checkmark  & & 72.41 \footnotesize{(\textcolor{blue}{10.76$\uparrow$})} & 77.29 \footnotesize{(\textcolor{blue}{3.15$\uparrow$})}\\
\checkmark & &\checkmark  & 74.33 \footnotesize{(\textcolor{blue}{12.68$\uparrow$})} & 77.44 \footnotesize{(\textcolor{blue}{3.30$\uparrow$})}\\
\checkmark &\checkmark & \checkmark & \textbf{76.17} \footnotesize{(\textcolor{blue}{14.52$\uparrow$})}& \textbf{78.76}\footnotesize{(\textcolor{blue}{4.62$\uparrow$})}\\
\bottomrule
\end{tabular}
% }
\caption{
Ablation studies on our AugSeg. 
% Results are reported on VOC and Cityscapes under the 1/4 partition protocols, both using ResNet-50 as the encoder.
``MT" means the standard mean-teacher semi-supervised framework. $\mathcal{A}_r$ and $\mathcal{A}_a$ represent the two main augmentation strategies, the random intensity-based and adaptive label-injecting augmentations, respectively. Improvements over the supervised baseline are highlighted in \textcolor{blue}{blue}.}
\label{tab:abl:modules}
\end{table}

\subsection{Comparison with SOTAs}
We demonstrate the superiority of AugSeg by comparing it with current SOTAs on both datasets under different partition protocols. Since U$^2$PL prioritizes selecting high-quality labels from classic VOCs for testing on blender VOC~\footnote{https://github.com/Haochen-Wang409/U2PL/issues/3}, we reproduce the supervised baseline and its performance on ResNet-50 for fair comparisons.

\textbf{Pascal VOC 2012}. In \Cref{tab:voc:fine} and \Cref{tab:voc:blender}, we compare our AugSeg with recent SOTAs on classic and blender VOC, respectively. It can be clearly seen from \Cref{tab:voc:fine} that, despite its simplicity, AugSeg can consistently outperform current SOTAs by a large margin, \eg obtaining a 4.45\% performance gain on 1/8 split using R101 as the encoder. Note that AugSeg can even achieve higher performance of 71.09\% with only 92 labels than the previous SOTA performance of 71.00\% with 183 labels. We can also observe that the performance gains become more noticeable and obvious when using ResNet-50 as the encoder and when fewer labels, \eg 92 and 182 labels, are available. Even though the performance gap between different SSS methods is decreasing as more labeled data is involved, our method can still improve the previous SOTA by 1.91\% and 1.35\% with 1/2 and full fine annotations, respectively.

As shown in \Cref{tab:voc:blender}, under 1/16, 1/8 and 1/4 partition protocols with the same split as U$^2$PL, our AugSeg obtains new SOTA performance of 79.29\%, 81.46\%, and 80.50\% based on ResNet-101, which obtains around 2\% performance improvements again previous SOTA. It is noteworthy that AugSeg obtains a higher mIoU with 1323 labels than that with 2646. It is simply because the 2646 split involves more noisy (coarsely-annotated) labels than 1323 split
(there are 1464 fine annotations in total). As discussed in \cref{sec:method:adaptive}, AugSeg can make full use of labeled data to stabilize the training on unlabeled data. Thus the noisy labeled information will degrade the performance of AugSeg. This is also why the superiority of AugSeg is more noticeable in \Cref{tab:voc:fine} than in \Cref{tab:voc:blender} with the same split as CPS~\cite{sss21cps}.

\textbf{Cityscapes}. In \Cref{tab:citys}, we evaluate our method on more challenging Cityscapes, using ResNet-50 and ResNet-101 as the encoder, respectively. We can easily see that AugSeg can readily outperform other SSS methods, especially with scarce labels. Though AugSeg is embarrassingly simpler than the recent SOTA U$^2$PL in terms of the training procedure and encoded feature size, AugSeg can improve U$^2$PL by 4.92\%, 3.45\%, 3.09\%, and 1.38\%, using ResNet-101 as the encoder, under 1/16, 1/8, 1/4, and 1/2 partition protocols, respectively. Not relying on advanced unsupervised techniques or multiple trainable models, AugSeg can consistently achieve the best performance on SSS benchmarks. Such impressive performance improvement further demonstrates the effectiveness and importance of our claim that various data augmentation should be simplified and adjusted to better adapt to semi-supervised learning.

In addition, we highlight the importance of labeled samples in semi-supervised learning, in terms of the quantity and quality. First, regardless of different semi-supervised approaches, we can see from Tables~\ref{tab:voc:fine}, ~\ref{tab:voc:blender} and ~\ref{tab:citys} that providing more labeled samples can easily boost the semi-supervised performance. Second, comparing the performance on classic and blended VOCs, we observe that the quality of labeled samples is always crucial. For example, our AugSeg can achieve a high performance of $78.80\%$ using 366 high-quality labels but require 2646 labels from the blender dataset to obtain a comparable mIOU of $78.82\%$.

\begin{figure}
    \centering
    \includegraphics[width=0.92\linewidth]{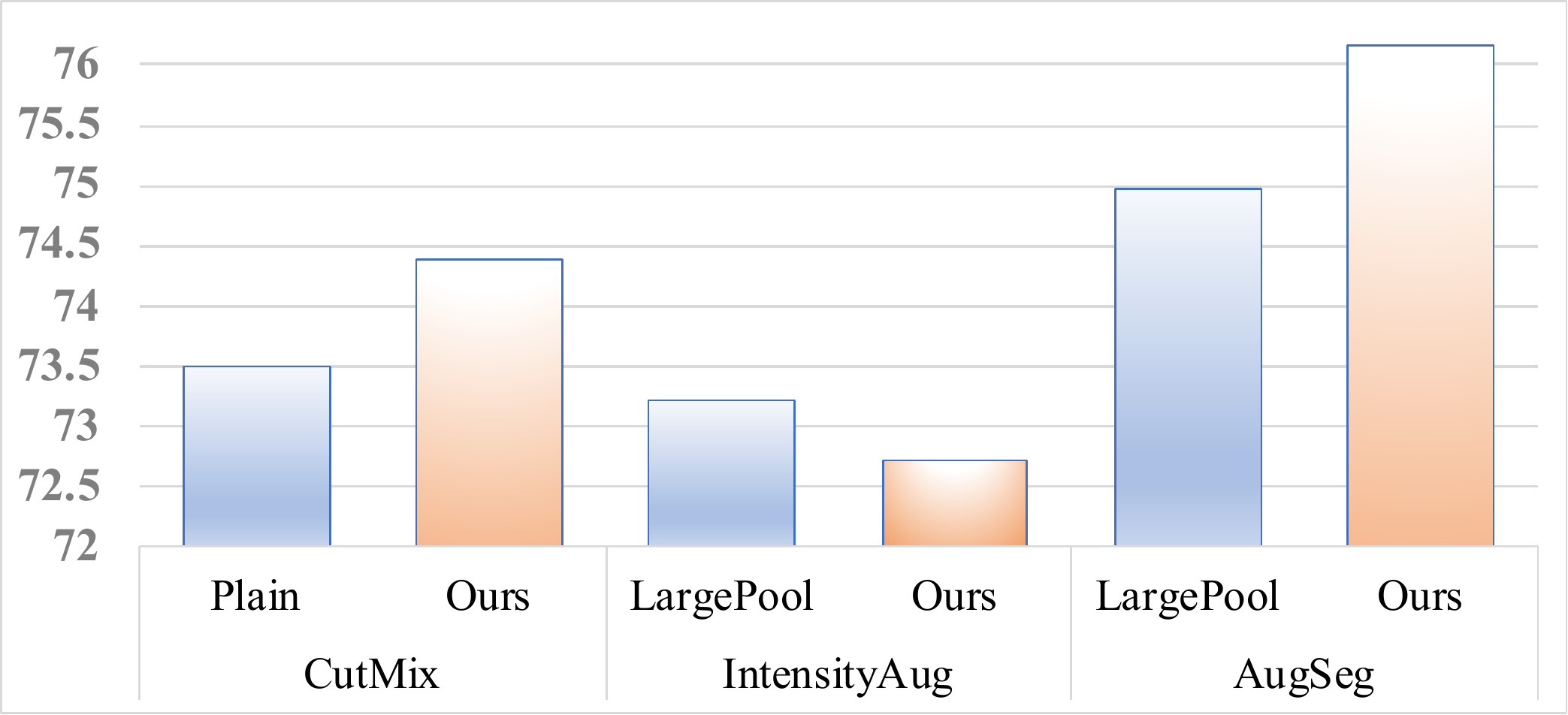}
    \caption{Ablation studies on different designs of AugSeg, where 
 ``LargePool" refers to the augmentation pool in \cite{sss21simple}.}
    \label{fig:abl:pool}
\end{figure}

\subsection{Ablations studies}
In this section, we conduct a series of ablations studies on Pascal VOC 2012 with 366 (1/4) labels and Cityscape with 744 (1/4) labels using ResNet-50 as the encoder.

\textbf{Effectiveness of different components of AugSeg}.  We first investigate the effectiveness of each component of AugSeg in \Cref{tab:abl:modules}. Our simplified augmentations $\mathcal{A}_r$ and $\mathcal{A}_a$ can effectively improve the SSS performance, obtaining 10.76\% and 12.68\% improvements against the supervised baseline, 3.35\% and 5.27\% improvements against the plain mean-teacher, on VOC 2012 with 366 labels, respectively. Using adaptive label-injecting augmentation $\mathcal{A}_a$ can obtain better performance than using $\mathcal{A}_r$ individually. 
Integrating both augmentations can further improve each individual component and achieve the best performance.

\begin{figure}
    \centering
    \includegraphics[width=0.9\linewidth]{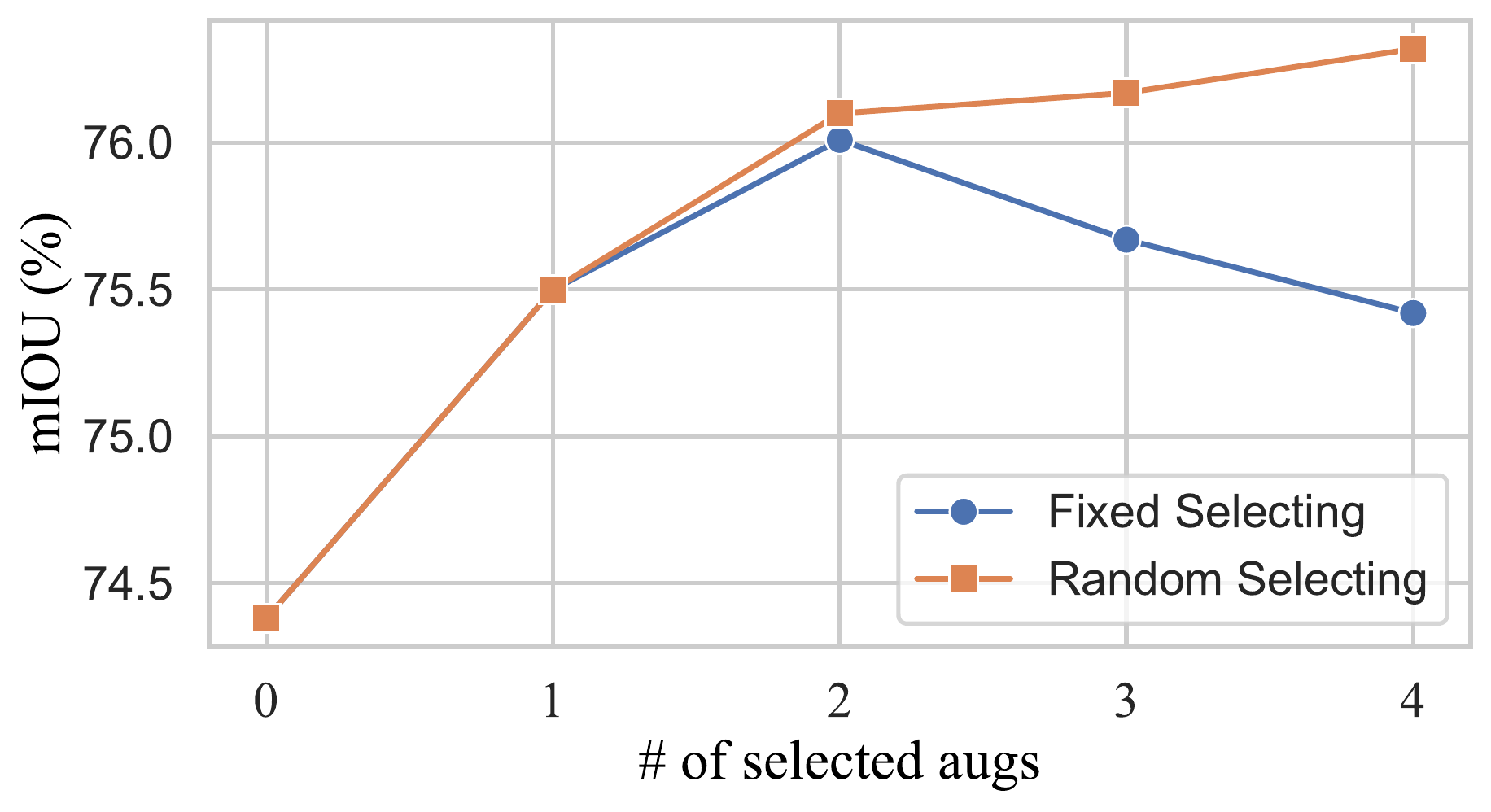}
    \caption{Impact of different selecting strategies in intensity-based augmentations with different numbers of selected operations.}
    \label{fig:abl:comp}
\end{figure}

\textbf{Impact of different augmentation designs}. As shown in \Cref{fig:abl:pool}, we test the impact of our simplified and adaptive designs on intensity-based and cutmix-based augmentations. We adopt $\mathcal{A}_a$ to stabilize the training on unlabeled data adaptively, which benefits the SSS training apparently. Following SimpleBaseline~\cite{sss21simple}, we also examine our random intensity-based designs with a larger augmentation pool. Consequently, although more augmentation selections can improve individual performance, incorporating both augmentations with more strong augmentation selections can degrade the performance, resulting from the discussed over-distortion issues in \cite{sss21simple}. Similar observations can be found from \Cref{fig:abl:comp} that using a fixed strategy to select more augmentations can harm the SSS performance. Our highly random and simplified designs can naturally alleviate this issue without introducing extra operations like Distribution-specific BN~\cite{chang2019domain}.

\begin{table}
    \centering
    \begin{tabular}{c|ccccc}
    \toprule
    $\lambda_u$ & 0.0 & 0.5 & 1.0 & 1.5& 2.0\\
    \midrule
    VOC (366) &  61.65 & 75.21 & 76.17 & 75.95 &\textbf{77.05} \\
    Citys (744) & 74.14 &  77.02 & 78.76 & \textbf{78.99} & 78.68 \\
    \bottomrule
    \end{tabular}
    \caption{Ablations on the loss weight $\lambda_u$, set as $1.0$ by default. }
    \label{tab:abl:weights}
\end{table}

\begin{table}
    \centering
    \begin{tabular}{c|ccccc}
    \toprule
    $k$ & 0 & 1 & 2 & 3 & 4\\
    \midrule
    VOC (366) &  74.38 & 75.50 & 76.10 & 76.17 & \textbf{76.32} \\
    Citys (186) & 71.26 & 72.10 & 73.42 & \textbf{73.73} & 73.03 \\
    Citys (744) & 77.44 & 78.34 & 78.11 & \textbf{78.76} & 78.48 \\
    \bottomrule
    \end{tabular}
    \caption{Ablations on the maximum number of selected intensity-based augmentations, using R50 as the encoder. $k=3$ by default.}
    \label{tab:abl:k}
\end{table}

\textbf{Impact of hyper-parameters}. We also examine the segmentation performance of AugSeg with different maximum numbers of selected augmentations, $k$ and different values of unsupervised loss weight $\lambda_u$, in \Cref{tab:abl:k} and \Cref{tab:abl:weights}, respectively. $\lambda_u\!=\!0$ means no training on unlabeled data while $k\!=\!0$ means not applying the rand intensity-based augmentation. We can observe that thanks to our random design, a larger $k$ will not degrade  but enhance the SSS performance, compared to the results with fixed selecting in \Cref{fig:abl:comp}.
It can also be seen from both tables that no specific parameters can consistently outperform others on both datasets. To keep our method simple and consistent, we set $k=3$ and $\lambda_u=1.0$ for all runs by default.

\begin{figure}
    \centering
    \includegraphics[width=0.999\linewidth]{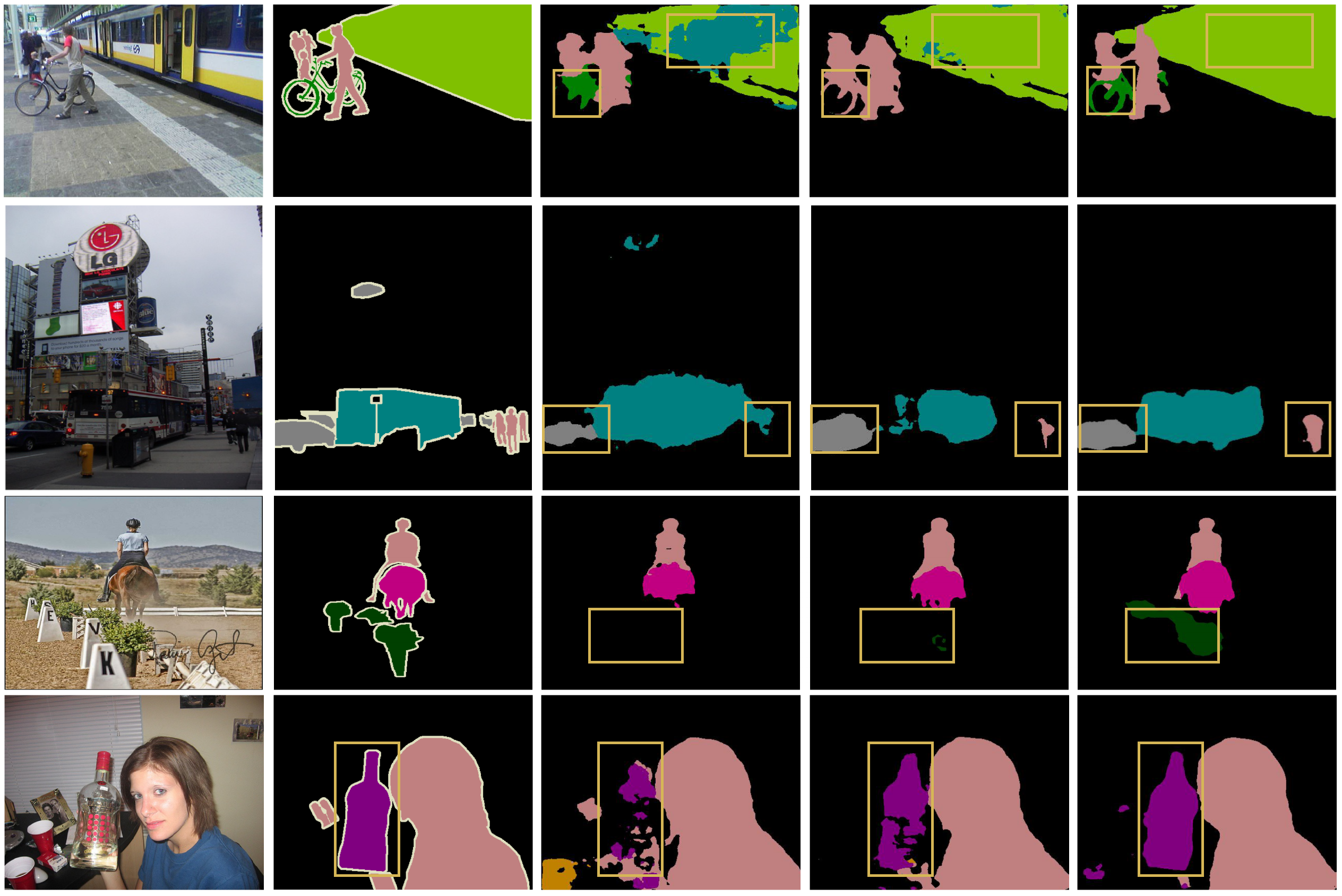}
    \caption{Qualitative results on Pascal VOC 2012 with 366 labels and ResNet-50 as the encoder. Columns from left to right denote the original images, the ground-truth, the supervised results, the plain MT results, and our AugSeg results, respectively. }
    \label{fig:abl:vis}
\end{figure}

\textbf{Qualitative Results}. \Cref{fig:abl:vis} shows some qualitative results on the Pascal VOC 2012 dataset. The supervised baseline obtains the worse segmentation results, \eg, not capable of differentiating the train and bus. Using more unlabeled data in the plain MT can enhance the model's capability to separate confusing classes. Among the comparisons, some challenging small-sized objects, such as the wheels, grass, or humans in a large background, can only be effectively identified by our AugSeg, which further demonstrates the effectiveness of AugSeg. Though our method obtains the SOTA SSS performance, AugSeg is \textbf{limited} at identifying some hard-to-segment objects, \eg, cars in the advertisement. We believe there is still great potential to further improve the SSS performance on top of our AugSeg.

%%%%%%%%%%%%%%%%%%%%%%%%%%%%%%%%%%%%
%%  5. Conclusion
%%%%%%%%%%%%%%%%%%%%%%%%%%%%%%%%%%%%

\section{Conclusion}

In this paper, we propose AugSeg, a simple-yet-effective approach to semi-supervised semantic segmentation. Unlike recent SSS studies that tend to combine increasingly complicated mechanisms, AugSeg follows a standard two-branch teacher-student framework to train models on labeled and unlabeled data jointly. The key to AugSeg lies in the simplification and revisions of two existing augmentation models, \ie, the random intensity-based and adaptive label-injecting CutMix-based augmentations. Without any additional complicated designs, AugSeg readily obtains new SOTA performance on popular SSS benchmarks under different partition protocols. We hope our AugSeg can serve as a strong baseline for future SSS studies.

%%%%%%%%% REFERENCES
{\small
\bibliographystyle{ieee_fullname}
\bibliography{augseg}
}

\end{document}